\documentclass[letterpaper]{article} % DO NOT CHANGE THIS
\usepackage{aaai2026}  % DO NOT CHANGE THIS
\usepackage{times}  % DO NOT CHANGE THIS
\usepackage{helvet}  % DO NOT CHANGE THIS
\usepackage{courier}  % DO NOT CHANGE THIS
\usepackage[hyphens]{url}  % DO NOT CHANGE THIS
\usepackage{graphicx} % DO NOT CHANGE THIS
\usepackage{natbib}  % DO NOT CHANGE THIS AND DO NOT ADD ANY OPTIONS TO IT
\usepackage{caption} % DO NOT CHANGE THIS AND DO NOT ADD ANY OPTIONS TO IT
\usepackage{algorithm}
\usepackage{algorithmic}

\usepackage{newfloat}
\usepackage{listings}
\DeclareCaptionStyle{ruled}{labelfont=normalfont,labelsep=colon,strut=off} % DO NOT CHANGE THIS
\floatstyle{ruled}
\newfloat{listing}{tb}{lst}{}
\floatname{listing}{Listing}
\usepackage{amsmath} 
\usepackage{amsfonts}
\usepackage{booktabs} 
\usepackage{xspace} 
\usepackage{multirow} 
\usepackage{pifont}
\usepackage{subcaption} 

\title{Correcting False Alarms from Unseen:\\ Adapting Graph Anomaly Detectors at Test Time}
\author {
    Junjun Pan\textsuperscript{\rm 1},
    Yixin Liu\textsuperscript{\rm 1},
    Chuan Zhou\textsuperscript{\rm 2},
    Fei Xiong\textsuperscript{\rm 3},
    Alan Wee-Chung Liew\textsuperscript{\rm 1},
    Shirui Pan\textsuperscript{\rm 1}\thanks{Corresponding  Author}
}
\affiliations {
    \textsuperscript{\rm 1}School of Information and Communication Technology, Griffith University, Queensland, Australia \\
    \textsuperscript{\rm 2}Academy of Mathematics and Systems Science, Chinese Academy of Sciences, Beijing, China\\
    \textsuperscript{\rm 3}School of Electronic and Information Engineering, Beijing Jiaotong University, Beijing, China\\
    \{junjun.pan, yixin.liu, a.liew, s.pan\}@griffith.edu.au, zhouchuan@amss.ac.cn, 
    xiongf@bjtu.edu.cn
}

\usepackage{bibentry}
\newcommand{\ourmethod}{TUNE\xspace}
\begin{document}

\maketitle

\begin{abstract}
Graph anomaly detection (GAD), which aims to detect outliers in graph-structured data, has received increasing research attention recently. However, existing GAD methods assume identical training and testing distributions, which is rarely valid in practice. In real-world scenarios, unseen but normal samples may emerge during deployment, leading to a \textbf{normality shift} that degrades the performance of GAD models trained on the original data. Through empirical analysis, we reveal that the degradation arises from (1) {semantic confusion}, where unseen normal samples are misinterpreted as anomalies due to their novel patterns, and (2) {aggregation contamination}, where the representations of seen normal nodes are distorted by unseen normals through message aggregation. While retraining or fine-tuning GAD models could be a potential solution to the above challenges, the high cost of model retraining and the difficulty of obtaining labeled data often render this approach impractical in real-world applications. To bridge the gap, we proposed a lightweight and plug-and-play \textbf{T}est-time adaptation framework for correcting \textbf{U}nseen \textbf{N}ormal patt\textbf{E}rns (\textbf{TUNE}) in GAD. To address semantic confusion, a graph aligner is employed to align the shifted data to the original one at the graph attribute level. Moreover, we utilize the minimization of representation-level shift as a supervision signal to train the aligner, which leverages the estimated aggregation contamination as a key indicator of normality shift. Extensive experiments on 10 real-world datasets demonstrate that TUNE significantly enhances the generalizability of pre-trained GAD models to both synthetic and real unseen normal patterns.
\end{abstract}



% Comment it out to save space
\begin{links}
    \link{Code}{https://github.com/CampanulaBells/TUNE}
    % \link{Datasets}{https://aaai.org/example/datasets}
    % \link{Extended version}{https://aaai.org/example/extended-version}
\end{links}

\section{Introduction}

Graph anomaly detection (GAD) aims to identify abnormal entities that deviate from the dominant patterns in a graph~\cite{ma2021comprehensive}. 
It has gained significant attention due to its critical role in safeguarding a wide range of graph-based applications, including social networks~\cite {bian2020rumor}, online services~\cite{wang2022wrongdoing}, and financial transaction networks~\cite{li2022internet}. 
In recent years, graph neural networks (GNNs)~\cite{liu2025graph,li2026assemble,li2024noise,shen2025understanding} have gained popularity in GAD due to their strong capabilities in capturing complex graph anomaly patterns. 
While several unsupervised GNN-based GAD methods can bypass the costly process of obtaining labeled anomalies~\cite{liu2021anomaly, ding2019deep, pan2023prem}, mainstream approaches still predominantly focus on supervised paradigms, which often yield higher detection accuracy in real-world GAD scenarios, such as fraud detection, intrusion detection, and misinformation spotting~\cite{tang2022rethinking, feng2025bimark, chen2024consistency}.

\begin{figure}[t]
  \centering
    \begin{subfigure}[t]{0.53\linewidth}
    \centering
    \includegraphics[width=\linewidth]{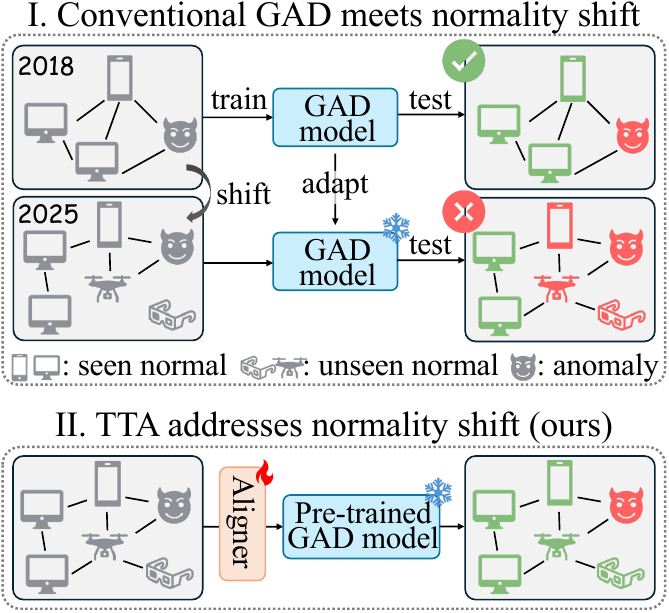}
    \caption{Sketch maps}\label{subfig:intro_sketch}
  \end{subfigure}
  \hfill
  \begin{subfigure}[t]{0.45\linewidth}     
    \centering
    \includegraphics[width=\linewidth]{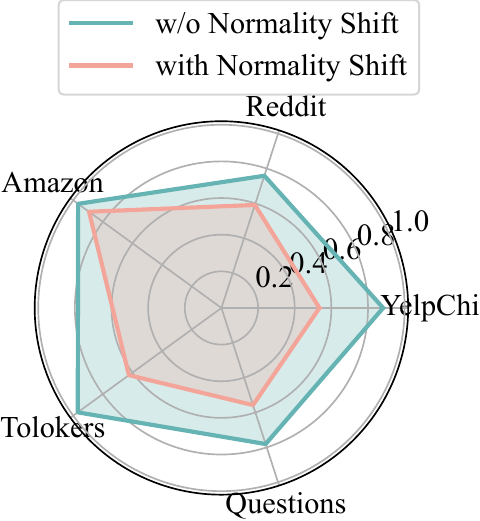}
    \caption{Performance drop}\label{subfig:intro_left}
  \end{subfigure}
\caption{(a) Sketch maps of conventional GAD methods under normality shift and our solution. (b) Performance drop of BWGNN under data with normality shift.}
\label{fig:intro}

\end{figure}

Despite the promising performance of existing GAD methods, their reliance on the assumption of identical training and testing data distributions may limit their performance in real-world scenarios, where new categories of normal entities can emerge and change the data distribution~\cite{baitieva2025beyond,tan2025bisecle,chen2025uncertainty,li2025clip}. Taking Figure~\ref{subfig:intro_sketch}-I as an example, in a product recommendation graph for electronics, the training nodes may previously consist of traditional devices such as laptops and smartphones, while at test time, new product categories like drones or smartglasses may emerge and become prevalent. We refer to this phenomenon as \textbf{normality shift}~\cite{han2023anomaly}, where the definition of ``normal'' evolves due to the emergence of previously unseen but benign categories. Given its ubiquity in real-world applications, this shift raises a fundamental question: \textit{Can GAD models pre-trained on the original normality distribution effectively adapt to new and evolving definitions of normality?}%Can the GAD models pre-trained on xxx effectively adapt to normality shift? 

To answer this question, we first conduct a preliminary experiment to examine how GAD models pre-trained on original data perform when directly adapted to data with unseen normal samples. As shown in Figure~\ref{subfig:intro_left}, the normality shift leads to a substantial performance drop on all datasets, highlighting the limitation of the generalizability of GAD models. To better understand the causes behind such performance degradation, we conduct a series of empirical analyses to investigate how unseen normal samples affect pre-trained GAD models. Our studies reveal a two-fold cause: \textit{on the one hand}, due to the unfamiliar patterns of the novel normal categories, the model suffers from \textbf{semantic confusion} between unseen normal samples and anomalies, leading to incorrect predictions on unseen normals; \textit{on the other hand}, the message propagation mechanism introduces \textbf{aggregation contamination}, where representations of seen normals are distorted by neighboring unseen normals, further degrading the model to distinguish known normal patterns. To sum up, the misclassification of both seen and unseen normal samples degrades the overall performance of GAD models under normality shift.

To address the above issues, a straightforward approach is to perform supervised fine-tuning or retrain the GAD models on the shifted data. 
Unfortunately, in real-world applications where data continuously evolves, frequently retraining the model from scratch is often impractical~\cite{ding2021cross}. Furthermore, identifying and labeling unseen normal categories is both costly and time-consuming, preventing us from fine-tuning GAD models on shifted data. To overcome these barriers, a more feasible solution is to leverage unsupervised test-time adaptation (TTA) techniques to adjust the shifted data for better alignment with the pre-trained GAD models through unsupervised objectives. 
Nevertheless, existing graph TTA methods~\cite{ju2023graphpatcher, jin2023empowering, mao2024source} are typically designed for node classification tasks and rely on several assumptions, such as graph homophily or balanced label distributions, to guide the adaptation. These assumptions do not hold in GAD settings, making it suboptimal to directly apply such methods. Moreover, these methods overlook the semantic confusion and aggregation contamination issues inherent in GAD with normality shift, thereby limiting their adaptation effectiveness.

To fill the gap, we introduce a \textbf{T}TA framework for correcting \textbf{U}nseen \textbf{N}ormal patt\textbf{E}rns (\ourmethod) in GAD, a plug-and-play lightweight method that can be integrated with diverse GAD model architectures without requiring any model retraining or fine-tuning. To deal with semantic confusion, we employ a graph aligner to correct the normality shift at the graph attribute level, which corrects the semantic deviation while preserving integration flexibility. For the aggregation contamination problem, rather than directly addressing it, we in turn leverage it to construct the supervision signals for unsupervised TTA. Specifically, \ourmethod employs an aggregation-free dual branch to simulate the undistorted node representations, and utilizes the divergence between the aggregation-free and aggregation-based branches to indicate the degree of aggregation contamination. Through minimizing this divergence, the framework optimizes the graph aligner, thereby reducing the impact of normality shift and hence enabling unsupervised TTA for GAD models. In summary, this paper makes the following contributions:

\noindent\textbf{Problem}. We identify the challenge of normality shift in real-world GAD scenarios and conduct comprehensive analyses to uncover the underlying difficulties, namely semantic confusion and aggregation contamination. 

\noindent\textbf{Methodology}. We introduce \ourmethod, a novel unsupervised TTA approach that can be integrated with any pre-trained GAD model in a plug-and-play manner to mitigate the impact of normality shift in GAD.

\noindent\textbf{Experiments}. We perform extensive experiments on 10 benchmark datasets, and the results demonstrate the effectiveness and generalizability of \ourmethod.

\section{Related Work}
In this section, we briefly review two related research directions, i.e., graph anomaly detection (GAD) and test-time adaptation (TTA). A more comprehensive literature review is available in Appendix A.

\noindent \textbf{Graph Anomaly Detection (GAD) } 
aims to identify outliers in graph-structured data. Unsupervised GAD methods typically identify anomalies via unsupervised learning techniques~\cite{ding2019deep,liu2021anomaly,liu2023towards,caileveraging,wugraph,zhao2025freegad}, while supervised GAD methods utilize annotations to build a more reliable anomaly classifier-based anomaly detector~\cite{dou2020enhancing,liu2024arc}. Existing methods either explicitly suppress it with carefully designed model architectures~\cite{gao2023addressing, shi2022h2, zhuo2024partitioning, yu2025dynamic,wu2025myopia}, or adopt spectral GNNs that are less sensitive to the local heterophilic connections~\cite{tang2022rethinking, gao2023addressing, xu2024revisiting, pan2025label}. Despite their performance, these methods assume a shared distribution between training and testing, limiting performance under normality shift~\cite{kim2024model}.

While recent works on open-set GAD~\cite{wang2023open, caileveraging} or cross-domain GAD~\cite{ding2021cross} focus on enhancing generalization to unseen domains~\cite{pan2025survey}, they require additional modules and training losses during the pre-training phase. This limits their effectiveness in adapting GAD models to normality shift.

\noindent \textbf{Graph Test Time Adaptation (TTA)} enables a pre-trained node classifier to adapt to distribution shift in unseen testing graphs, without fine-tuning the model itself. GTrans~\cite{jin2023empowering} pioneers the use of a data-centric graph TTA that can be seamlessly integrated with any model architecture, which is trained with a contrastive learning paradigm~\cite{deng2025THESAURUS,fu2025less}.
Similarly, GraphPatcher~\cite{ju2023graphpatcher} mitigates degree bias by iteratively generating virtual nodes to repair corrupted low-degree graphs.  With the assumptions on graph homophily and label balance, SOGA~\cite{mao2024source} directly fine-tunes the pre-trained model by enforcing structure consistency and maximizing mutual information between the input and output. Despite their promising results in classification tasks, they are not directly applicable to GAD due to challenges such as strong label imbalance and the camouflage effect~\cite{dou2020enhancing}.

\section{Problem Definition}
\noindent \textbf{Notations.} 
Let $\mathcal{G} = (\mathcal{V}, \mathcal{E}, \mathbf{X})$ denote an attributed graph with $n$ nodes and $m$ edges, where $\mathcal{V} = \{v_1, ..., v_n\}$ is the node set, $\mathcal{E}$ is the edge set, and $\mathbf{X} = \{\mathbf{x}_1, ..., \mathbf{x}_n\}$ is the node feature matrix, with $\mathbf{x}_i \in \mathbb{R}^d$ representing the feature vector of node $v_i$. The adjacency matrix is denoted by $\mathbf{A} \in \mathbb{R}^{n \times n}$, where $\mathbf{A}_{ij} = 1$ if $(v_i, v_j) \in \mathcal{E}$ and $0$ otherwise.

\noindent \textbf{Conventional GAD Problem. }
GAD aims to classify anomalous nodes $\mathcal{V}_a$  from normal nodes $\mathcal{V}_n = \mathcal{V} \setminus \mathcal{V}_a$. This is achieved by learning an anomaly scoring function $f: \mathcal{V}\rightarrow \mathbb{R}$ that predicts anomaly score $s_i$ for each node $v_i$, where a higher score indicates a higher likelihood of being anomalous. In the conventional setting,  it is assumed that the training graph $\mathcal{G}_\text{train}$ and the test graph $\mathcal{G}_\text{test}$ are drawn from the same underlying distribution,  i.e.,  $\mathcal{G}_\text{train} \sim \mathbb{P}$ and $\mathcal{G}_\text{test} \sim \mathbb{P}$. This assumption enables the scoring function $f$ learned on  $\mathcal{G}_\text{train}$ to generalize effectively to $\mathcal{G}_\text{test}$.

\noindent\textbf{GAD with Test-Time Normality Shift. }
Formally, let the training and testing graphs be denoted as $\mathcal{G}_\text{train}$ and $\mathcal{G}_\text{test}$ respectively. The normality shift problem in GAD can be defined as the scenario where the underlying distribution $\mathbb{P}$ is non-stationary between training and testing~\cite{kim2024model}, i.e., $\mathbb{P}_\text{train} \neq \mathbb{P}_\text{test}$, due to the emergence of previously unseen normal nodes $\mathcal{V}_\text{un} \subset \mathcal{V}_\text{test}$. These unseen normal nodes can introduce novel and semantically valid patterns into the test graph that were absent during training, resulting in a distribution shift that leads existing GAD models to mistakenly assign them a high anomaly score, as illustrated in Figure~\ref{subfig:intro_left}. 

\section{Motivation and Analysis}
\begin{figure}[t]
  \centering
  \begin{subfigure}[b]{0.48\linewidth}
    \includegraphics[width=\linewidth]{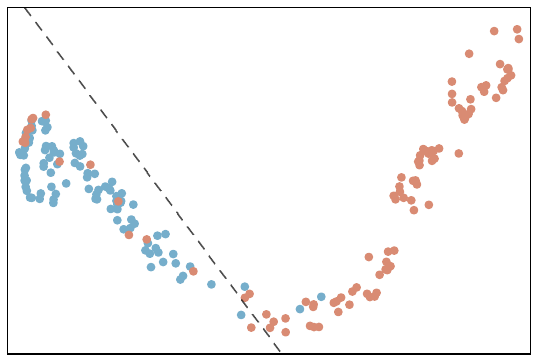}
    \caption{Without unseen normals}
    \label{subfig:before}
  \end{subfigure}
  \hfill
  \begin{subfigure}[b]{0.48\linewidth}
    \includegraphics[width=\linewidth]{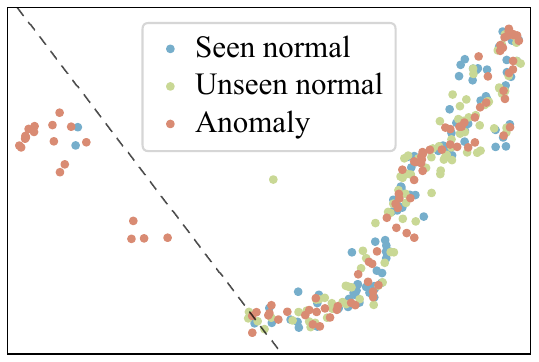}
    \caption{With unseen normals}
    \label{subfig:after}
  \end{subfigure}

  \vskip\baselineskip

  \begin{subfigure}[b]{0.98\linewidth}
    \includegraphics[width=\linewidth]{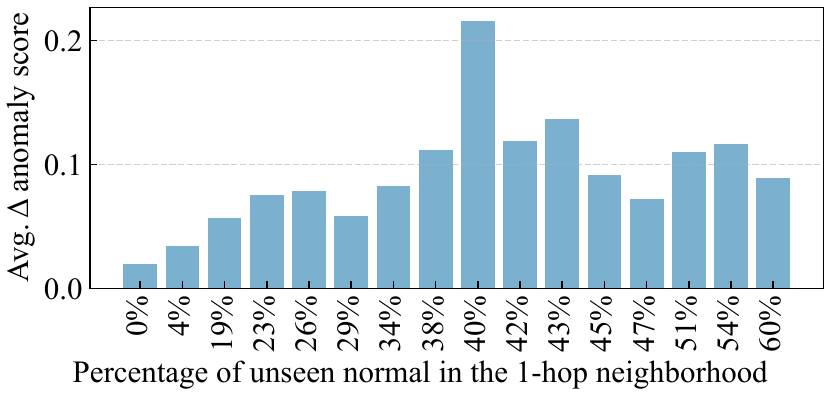}
    \caption{Anomaly score change vs. unseen normals in 1-hop neighbors}
    \label{subfig:delta_score}
  \end{subfigure}

  \caption{Motivate experiments on BWGNN.}
  \label{fig:preliminary}
\end{figure}

Normality shift frequently occurs in real-world applications such as e-commerce, driven by the continuous emergence of new products. In order to understand how normality shift affects GAD models, in this section, we expose the impact of normality shifts on the effectiveness of GAD models pre-trained on original data. Specifically, we first examine the generalization capability of pre-trained GAD models on unseen normal samples and find that \textbf{semantic confusion} between unseen normals and anomalies is a key factor that leads to generalization failure. 
Furthermore, we observe that the presence of unseen normal instances also affects the identification of previously seen normal data, caused by the \textbf{aggregation contamination} phenomenon. We provide empirical analysis and discussion to explain these issues.

\subsection{Semantic Confusion of Unseen Normals}
In existing GAD methods, a common assumption is that the training and testing sets are drawn from the same distribution. 
Yet in real-world scenarios, new but benign node categories often emerge and follow valid behavioral patterns. Although these instances may appear novel to the model, they should not be treated as anomalies.
This raises a key question that we seek to explore: \textit{Can GAD methods effectively generalize to unseen normal categories?} 
To answer this question, we evaluate the generalizability of BWGNN~\cite{tang2022rethinking} on five real-world GAD datasets to assess how the pre-trained model is affected by unseen normal categories introduced in the test split. As shown in Figure~\ref{subfig:intro_left}, we observe a consistent and significant performance drop on datasets from different domains, showing that the presence of unseen normal classes severely reduces performance. 

\noindent\textbf{Analysis:} From the perspective of unseen normal samples, we attribute the performance degradation to the \textbf{semantic confusion} between new normals and anomalies. Specifically, GAD models are typically trained to assign low anomaly scores to seen normal instances and high scores to anomalies, based on the observed training distribution. However, due to the inherent sparsity of anomalies and the limited diversity of normal categories, the learned pattern of normality tends to be overly specific. Consequently, when novel but benign categories emerge during testing, their unfamiliar patterns often lead the model to assign inflated anomaly scores. This indicates that pre-trained GAD models tend to overfit to seen normal patterns and struggle to generalize to unseen ones.

\noindent\textbf{Empirical Discussion:} To investigate how unseen normal samples are misclassified by the model, we visualize the learned representations and the corresponding decision boundary on Tolokers dataset~\cite{platonov2023critical}, comparing the model behavior without (Figure~\ref{subfig:before}) and with (Figure~\ref{subfig:after}) unseen normal samples. In Figure~\ref{subfig:before}, we observe that the seen normal samples (in blue) and anomalies (in red) are well-separated, indicating the reliable performance of the GAD model on data without normality shift. Nevertheless, in Figure~\ref{subfig:after}, the representations of unseen normal nodes (in green) are found to overlap significantly with those of anomalies, revealing that the model struggles to differentiate novel normal patterns from truly abnormal ones. The findings highlight that the semantic confusion between unseen normals and anomalies severely impairs the discrimination ability of GAD models under normality shift.

\subsection{Aggregation Contamination on Seen Normals}

According to Figure~\ref{subfig:after}, we not only observe semantic confusion between the unseen normal samples and anomalies, but also an unexpected effect: the representations of seen normal nodes (in blue), which were previously well separated from the anomalies in Figure~\ref{subfig:before}, also undergo a noticeable shift. Such a shift, in turn, degrades the model performance in distinguishing anomalies from previously learned normal patterns. This effect brings forth a puzzling question: \textit{Since the attributes of seen normal samples usually remain unchanged, why do their representations shift in the presence of unseen normal samples?}

\noindent\textbf{Analysis:} We suggest that this effect stems from \textbf{aggregation contamination}, where the representations of seen normal nodes are distorted due to the presence of unseen normals. This effect arises from the message aggregation mechanism in graph neural networks, where a node updates its representation by integrating information from its neighbors~\cite{wu2020comprehensive}. Consequently, unseen normals can influence nearby seen normals through aggregation, causing their representations to deviate from expected normal patterns. 
Despite several GAD methods employing techniques like spectral GNN~\cite{tang2022rethinking} or weighted aggregation~\cite{shi2022h2} to address similar issues caused by anomalies during message aggregation, these approaches fall short when dealing with unseen normals. A key distinction is that unseen normals emerge only at test time, leaving the model unable to learn how to suppress their influence.  Moreover, unlike anomalies that often exhibit distinct or extreme behaviors, unseen normals tend to follow partially familiar patterns, making their interference more subtle and thus harder to identify and mitigate.

\noindent\textbf{Empirical Discussion:} To assess the aggregation contamination effect, we examine how the presence of unseen normals alters the anomaly scores of seen normal nodes. Specifically, we group nodes based on the proportion of unseen normals in their 1-hop neighborhoods and compute the average change in anomaly score before and after introducing these unseen nodes. As shown in Figure~\ref{subfig:delta_score}, the presence of unseen normals consistently leads to increased anomaly scores for seen normal nodes. Moreover, the magnitude of this increase grows with a higher proportion of novel normal neighbors, supporting our hypothesis that message aggregation amplifies the effect. Notably, even nodes with no unseen normals in their 1-hop neighborhoods experience elevated anomaly scores, suggesting that the influence can propagate through multi-hop interactions. This confirms that aggregation from unseen normals can distort the representation of seen normals, leading to incorrect abnormality prediction.

\section{Methodology}
\begin{figure*}
  \centering
  \includegraphics[width=1\linewidth]{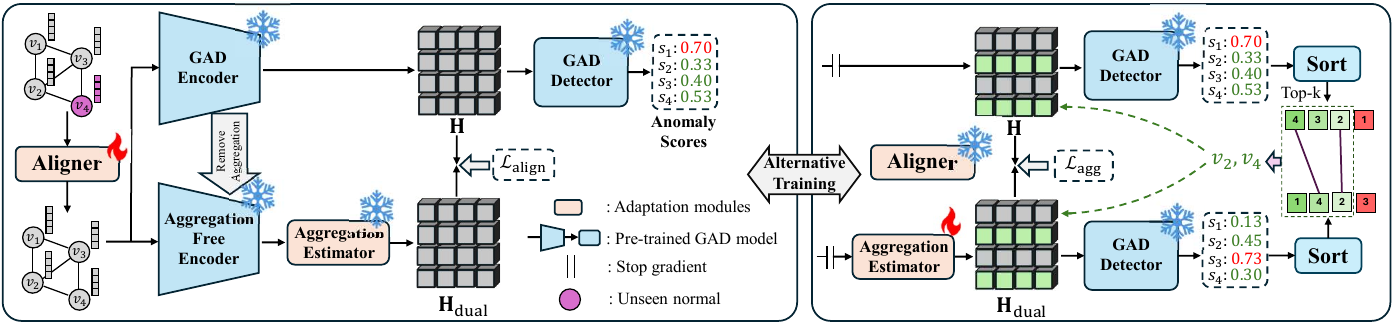}
  \caption{Overall framework of \ourmethod. \ourmethod addresses normality shift by leveraging a graph aligner and a dual-branch architecture. It captures aggregation contamination caused by unseen normals by measuring the discrepancy between representations from a main branch and an aggregation-free auxiliary branch. To ensure that the auxiliary branch provides contamination-free representations, an aggregation estimator is jointly trained with the aligner in an alternating manner, using high-confidence normal nodes. }
  \label{fig:method:architecture}
\end{figure*}

Our empirical study demonstrates that normality shift hinders GAD performance in real-world scenarios by causing both \textit{semantic confusion of unseen normals} and aggregation \textit{contamination on seen normals}. To address these challenges while minimizing deployment costs, we propose \ourmethod, an unsupervised test-time adaptation framework that can be seamlessly integrated with various pre-trained GAD models to enable generalization to unseen normals, without requiring supervised fine-tuning. As illustrated in Figure~\ref{fig:method:architecture}, \ourmethod employs a graph aligner to transform node features, alleviating the semantic confusion phenomenon by reducing the distribution shift of unseen normals. 
To train the aligner without labels, we design an alignment loss that leverages aggregation contamination as an indicator of normality shift, and then minimize the discrepancy between aggregated and uncontaminated node representations.

\subsection{Feature Transformation-based Graph Aligner}

To overcome the semantic confusion caused by normality shift in real-world applications, test-time adaptation offers a promising solution by aligning the feature distributions of unseen normal samples with those observed during training, while preserving the GAD models pre-trained on original data unchanged. Nevertheless, the diverse architectures of pre-trained GAD models make it challenging to design adaptation strategies that generalize across different GAD architecture designs. To this end, we introduce a simple yet effective graph aligner that enables data-centric test-time adaptation based on feature transformation. This design allows \ourmethod to be compatible with diverse GAD models without requiring supervised fine-tuning or architectural modifications.

In essence, normality shift arises because unseen normal nodes exhibit attribute patterns that deviate from those encountered during training. To address this, we project the features of unseen normals toward the distribution of seen normals, which realizes a model-agnostic adaptation. Specifically, we model this deviation as a learnable residual and train a multilayer perceptron (MLP) to estimate it based on the node feature matrix $\mathbf{X}$. The estimated shift $\Delta\mathbf{X}$ is then applied to produce the aligned features $\mathbf{X}'$:
\begin{equation}
\mathbf{X}'=f_\text{align}(\mathbf{X}; \theta)=\mathbf{X}+\Delta\mathbf{X}, \quad \Delta\mathbf{X}=\text{MLP}_\theta(\mathbf{X}).
\end{equation}
The aligned features are subsequently passed into the pre-trained GAD model to produce anomaly scores $\textbf{s}_\text{gnn}$:
\begin{equation}
   \mathbf{H} = f_{\text{enc}}(\mathbf{A}, \mathbf{X}'),  \quad \mathbf{s}_\text{test} = f_\text{det}(\mathbf{H} ),
\end{equation}
where $f_{\text{enc}}(\cdot)$ and $f_{\text{det}}(\cdot)$ denote the encoder and detector components of the pre-trained GAD model, respectively. 

To optimize the parameters $\theta$ of the graph aligner, an ideal learning objective is to maximize the supervised detection accuracy on the aligned test graph:
\begin{equation}
   \underset{\theta}{\text{argmin }} \mathcal{L}(\textbf{s}_\text{test}, \mathbf{y}_\text{test}),
\end{equation}
where $\mathcal{L}(\cdot,\cdot)$ denotes the supervised anomaly detection loss. This formulation guides the aligner to reduce the impact of normality shift by minimizing its negative impacts on downstream GAD performance, which can handle the semantic confusion issue. However, since the test-time ground-truth labels $\mathbf{y}_\text{test}$ are usually hard to acquire in the context of GAD, directly optimizing for detection performance is infeasible. Motivated by our empirical findings, we instead leverage aggregation contamination as an implicit indicator to estimate and mitigate the influence of normality shift in an unsupervised setting. 

\subsection{Aggregation Contamination-Guided Alignment}
Our empirical analysis demonstrates that unseen normals can distort the embedding of seen normal nodes through message aggregation, leading to elevated anomaly scores. If we could obtain representations that are not affected by such contamination, the normality shift could be estimated by comparing the contaminated representations with their uncontaminated counterparts, thereby guiding the optimization of the graph aligner. In other words, we can optimize the aligner by minimizing the representation shift caused by contaminated aggregation, which uses the contamination itself as feedback to counteract its effect.

To obtain contamination-free representations without label supervision, we construct an aggregation-free auxiliary branch from the GNN backbone in GAD models, where message aggregation is explicitly removed. In this way, we thereby isolate node representations from the influence of unseen normals. Specifically, the auxiliary branch replicates the encoder of the pre-trained GAD model, but removes message passing and aggregation, yielding an encoder $f_{\text{dual}}(\cdot)$ that generates node representations solely from ego features. This adjustment can be applied to a wide range of GNNs by replacing their aggregation operators with identity matrices. To further compensate for the absence of contextual signals, we incorporate an aggregation estimator $g(\cdot)$ that reconstructs aggregated embeddings $\mathbf{H}_{\text{dual}}$ using only ego representations, i.e.: 
\begin{equation}
    \mathbf{H}_{\text{dual}} = g(f_{\text{dual}}(\mathbf{X'})).
\end{equation}

Assuming the estimator $g(\cdot)$ can accurately approximate clean embeddings, the divergence between the outputs of the main branch and auxiliary branch reflects the impact of aggregation contamination. This signal is then used to optimize the graph aligner via an alignment loss:
\begin{equation}\label{eq:l_align}
    \mathcal{L}_{\text{align}} = \text{KLD}(\mathbf{H}|\mathbf{H}_{\text{dual}}),
\end{equation}
where $\mathbf{H}$ denotes the contaminated embedding from the main branch, and KLD is the Kullback–Leibler divergence. Minimizing this loss encourages the aligner to reduce the influence of normality shift in the input features, providing an effective unsupervised adaptation strategy across GAD architectures to handle shifted normal patterns.

\subsection{Training Aggregation Estimator} 
To reduce the influence of aggregation contamination on seen normals, our auxiliary branch encodes each node using only its ego features. However, while this eliminates contamination from unseen normals, it also discards potentially informative signals from clean ones. To compensate for the missing information, we introduce an aggregation estimator $g(\cdot)$ to reconstruct the aggregated representations from the aggregation-free ones.

\begin{table*}[t!]
\centering
\setlength{\tabcolsep}{1mm}
\resizebox{\textwidth}{!}{%
\begin{tabular}{l|cccccccc}
\toprule
\textbf{Model} & \textbf{Amazon} & \textbf{YelpChi} & \textbf{Reddit} & \textbf{Weibo} & \textbf{Tolokers} & \textbf{Questions} & \textbf{T-Finance} & \textbf{T-Social} \\
\midrule
CareGNN & 77.03±2.18 & 53.47±1.53 & 56.79±0.91 & 66.85±0.91 & 65.74±0.20 & 50.45±0.04 & 86.14±2.91 & OOM \\
H2-Fdetector & 61.19±13.84 & 53.34±0.75 & 51.31±3.58 & 61.56±12.08 & 61.30±2.21 & 50.44±4.08 & OOM & OOM \\
GAGA & 87.44±1.26 & 52.03±0.64 & 52.25±4.54 & 94.43±1.76 & 60.53±2.40 & 51.80±1.34 & 93.64±0.89 & OOT \\
PMP & 80.78±2.17 & 55.62±2.42 & 56.31±0.50 & 80.25±1.59 & \textbf{66.87±1.03} & 58.14±4.93 & 80.67±6.69 & OOM \\
ConsisGAD & 90.32±0.73 & 50.97±0.42 & 62.93±1.21 & 93.43±0.52 & 59.55±1.28 & 48.35±0.83 & 85.57±0.79 & 70.36±4.40 \\
\midrule
GCN & 21.21±1.02 & 47.70±0.16 & 62.88±0.44 & 86.30±14.06 & 63.20±0.26 & 51.58±0.44 & 84.08±1.32 & 69.79±8.03 \\
\quad + GTrans & 49.28±11.12 & 49.37±0.60 & 47.13±4.64 & 82.56±4.67 & 60.15±4.17 & 40.97±0.99 & 68.89±1.87 & OOM \\
\quad + GraphPatcher & OOM & OOM & 62.27±1.32 & 73.71±20.73 & 64.97±0.42 & 57.37±1.52 & OOM & OOM \\
\quad + SOGA & 22.81±2.90 & 49.30±0.32 & 61.08±0.65 & 79.17±20.35 & 64.67±2.01 & 50.91±0.45 & 77.18±2.07 & OOM \\
\quad + Ours & \underline{55.10±22.83} & \underline{54.50±0.74} &  \underline{\textbf{63.62±0.03}} & \underline{\textbf{96.16±0.01}} & \underline{66.27±0.07} & \underline{60.61±0.97} & \underline{85.53±0.04} &\underline{ 81.89±0.02} \\
\midrule
BWGNN & 83.38±1.75 & 51.87±0.71 & 59.62±1.50 & 93.04±1.10 & 62.02±0.87 & 50.40±0.45 & 92.35±1.27 & 80.14±2.08 \\
\quad + GTrans & 76.85±3.50 & 51.41±0.55 & 52.92±0.88 & 88.43±3.04 & 60.06±2.51 & 49.05±1.27 & 84.07±1.71 & OOM \\
\quad + GraphPatcher & OOM & OOM & 60.52±1.78 & 84.83±12.48 & 58.72±1.51 & 58.67±0.50 & OOM & OOM \\
\quad + SOGA & 77.61±2.60 & OOM & 58.34±2.47 & 77.21±5.65 & 62.26±1.21 & 49.68±0.41 & 85.12±2.84 & OOM \\
\quad + Ours & \underline{\textbf{92.19±0.67}} & \underline{\textbf{60.58±0.87}} & \underline{61.34±0.20} &\underline{ 95.26±0.49} & \underline{65.51±0.04} & \underline{\textbf{63.21±0.37}} & \underline{\textbf{94.28±0.53}} &\underline{82.72±0.01} \\
\midrule
GHRN & 81.58±2.19 & 52.75±0.35 & 61.17±1.01 & 90.34±2.48 & 60.34±0.59 & 49.65±0.82 & 92.72±0.82 & 85.73±1.94 \\
\quad + GTrans & 79.12±4.03 & 52.05±0.40 & 51.05±2.86 & 88.14±5.02 & 63.44±1.74 & 48.75±0.78 & 78.52±7.40 & OOM \\
\quad + GraphPatcher & OOM & OOM & 59.57±4.49 & OOM & 62.67±2.76 & 58.68±0.74 & OOM & OOM \\
\quad + SOGA & 77.62±1.57 & 52.36±0.49 & 56.87±2.50 & 77.26±7.22 & 62.67±1.08 & 48.57±0.87 & 87.50±4.03 & OOM \\
\quad + Ours & \underline{82.11±6.71} & \underline{58.36±0.66} & \underline{63.46±0.10 }&\underline{ 92.31±0.01} & \underline{66.17±0.42} & \underline{58.88±0.39} & \underline{93.28±0.04} & \underline{\textbf{88.42±0.01}} \\
\bottomrule
\end{tabular}
}

\caption{Performance (AUROC in \%) with standard deviation of all models on GFD datasets. \textbf{Bold} indicates the best overall performance across all models, while  \underline{underline} highlights the best result among TTA variants for each base model.}
\label{table:mainresults}
\end{table*}

To avoid reintroducing contamination during reconstruction, the estimator is trained using a carefully selected set of clean nodes with high-confidence normality. As illustrated in right of Figure~\ref{fig:method:architecture}, we obtain anomaly scores from two branches, $\textbf{s}$ and $\textbf{s}_{\text{dual}}$, by feeding 
%the main and auxiliary embeddings  偷空间
$\mathbf{H}$ and $\mathbf{H}_{\text{dual}}$ into the GAD detector. We then select nodes that rank in the top-$k$ percentile of normal confidence under both scores as training samples. A high confidence score under the auxiliary branch suggests the ego features of a node are normal, while a high score under the main branch implies its normality is not changed due to unseen normals, together indicating a clean supervision signal. The corresponding embeddings, $\mathbf{H}^{\text{top-}k}$ and $\mathbf{H}_{\text{dual}}^{\text{top-}k}$, are used to optimize the aggregation estimator $g(\cdot)$ via another KL-divergence loss:
\begin{equation}\label{eq:l_agg}
    \mathcal{L}_{\textit{agg}} = \text{KLD}(\mathbf{H}
    ^\text{top-k}|\mathbf{H}_{\text{dual}}^\text{top-k}),
\end{equation}
where $g(\cdot)$ is implemented as a linear transformation. Note that $\mathcal{L}_{\textit{agg}}$ is only used to update the aggregation estimator rather than the graph aligner. 
By supervising the estimator with high-confidence normal nodes, we ensure the reconstructed representations can be sufficient to approximate contamination-free aggregation, thereby improving the detection of normality shift.

\subsection{Overall Algorithm}

There are two learnable modules in the \ourmethod pipeline, graph aligner $f_\text{align}(\cdot)$ and aggregation estimator $g(\cdot)$. The optimization of these two modules is tightly coupled: while $g(\cdot)$ plays a critical role in the optimization objective of $f_\text{align}(\cdot)$ (i.e., Eq.~\eqref{eq:l_align}), $f_\text{align}(\cdot)$ also serves as the feature transformation input to $g(\cdot)$. In this case, to ensure the stable optimization of \ourmethod, we adopt an \textbf{alternative training} strategy. {Concretely}, we first fix the parameters of the aggregation estimator $g(\cdot)$ and optimize the graph aligner $f_{\text{align}}(\cdot)$ using the alignment loss in Eq.~\eqref{eq:l_align} (left part of Figure~\ref{fig:method:architecture}). Then, with $f_{\text{align}}(\cdot)$ frozen, we update $g(\cdot)$ under the supervision of Eq.~\eqref{eq:l_agg} (right part of Figure~\ref{fig:method:architecture}). These two steps are alternated iteratively until the training converges. The alternative training strategy results in an overall complexity of $\mathcal{O}(n + m)$. The algorithmic description and detailed complexity analysis are provided in Appendix B.

\section{Experiments}

\subsection{Experimental Settings}
\noindent \textbf{Datasets. } We conduct experiments on eight public GAD datasets from different domains, including Amazon~\cite{mcauley2013amateurs}, YelpChi~\cite{rayana2015collective}, Reddit~\cite{kumar2019predicting}, Weibo~\cite{kumar2019predicting}, Tolokers~\cite{platonov2023critical}, Questions~\cite{platonov2023critical}, T-Finance~\cite{tang2022rethinking}, and T-Social~\cite{tang2022rethinking}. Since these datasets only contain binary labels and do not explicitly define subtypes within the normal class, we apply clustering on the features of normal nodes and select the smallest cluster as unseen normals. To simulate a normality shift, the unseen normals are excluded from the training and validation sets. To further evaluate generalizability against natural unseen normals, we include two imbalanced node classification datasets, Photo and Computers~\cite{shchur2018pitfalls}. We treat classes with fewer than 5\% of the total nodes as anomalies, while the largest normal class is utilized as the unseen normal. More details are given in Appendix C. 

\noindent \textbf{Baseline and Evaluation Metrics. } We compare our approach against five state-of-the-art supervised GAD methods: CareGNN~\cite{dou2020enhancing}, H2-FDetector\cite{shi2022h2}, GAGA~\cite{xu2024revisiting}, PMP~\cite{zhuo2024partitioning}, and ConsisGAD~\cite{yu2025dynamic}. Moreover, we compare \ourmethod with three graph TTA baselines, including GTrans~\cite{jin2023empowering}, GraphPatcher~\cite{ju2023graphpatcher}, and SOGA~\cite{mao2024source}. Furthermore, we evaluate all TTA methods alongside \ourmethod with three popular GNN/GAD methods as backbones, including GCN~\cite{kipf2016semi}, BWGNN~\cite{tang2022rethinking}, and GHRN~\cite{gao2023addressing}. To simplify the implication, we adapt the implementation provided by GADBench~\cite{tang2023gadbench}. We utilize the suggested hyperparameters in their paper for all baselines. Similar to previous works~\cite{tang2023gadbench}, we employ AUROC and AUPRC (results see Appendix E) as evaluation metrics. We run each experiment 5 times and report the average and standard deviation. More experimental details are given in Appendix D. 

\subsection{Experimental Results}

\noindent\textbf{Performance Comparison.} The comparison results on eight real-world GFD datasets with constructed unseen normals are reported in Table~\ref{table:mainresults}. From the table, we can see that: \ding{182}~\ourmethod consistently and significantly improves performance on the three GNN backbones across all datasets, and outperforms SOTA methods in most datasets except on Tolokers. These results demonstrate the effectiveness of \ourmethod in diverse real-world GFD datasets. \ding{183}~While existing Graph TTA methods (GTrans, GraphPatcher, and SOGA) exhibit solid performance on classification tasks, they failed to achieve positive transfer in GAD with normality shift. This is mainly due to the unique challenges in GAD, such as severe label imbalance and heterophilic structures, which violate the core assumptions of these methods~\cite{mao2024source, FengIJCNN}. 
\ding{184}~The scalability of TTA methods is hindered by out-of-memory (OOM) issues on large-scale real-world GFD datasets such as T-Finance and T-Social. In contrast, \ourmethod demonstrates better scalability, underscoring its effectiveness in handling large-scale GAD applications like e-commerce and social networks.

\begin{figure}[t]
  \centering
  \begin{subfigure}[t]{0.49\linewidth}     
    \centering
    \includegraphics[width=\linewidth]{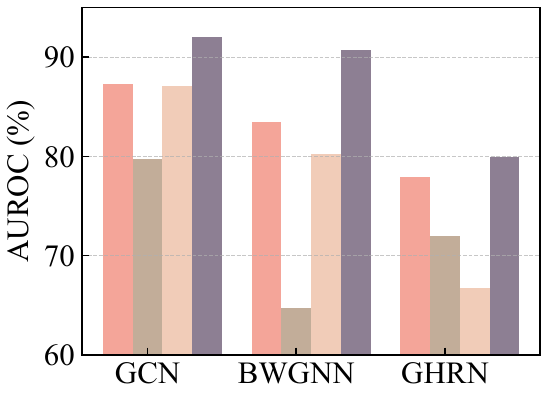}
    \caption{Photo}\label{subfig:bar_photo}
  \end{subfigure}
  \hfill
  \begin{subfigure}[t]{0.49\linewidth}
    \centering
    \includegraphics[width=\linewidth]{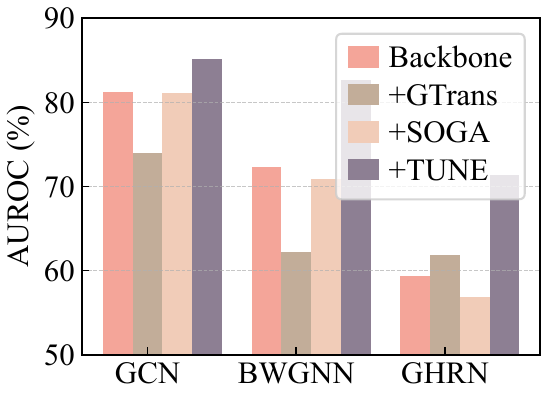}
    \caption{Computers}\label{subfig:bar_comp}
  \end{subfigure}
\caption{Performance on Photo and Computers datasets.}
\label{fig:realNN}
\end{figure}

\noindent \textbf{Real-World Unseen Normals. } To evaluate the effectiveness of our method under normality shift introduced by real-world unseen normal categories, we test it on two imbalanced classification datasets: Photo and Computer. As shown in Figure~\ref{fig:realNN}, while other Graph TTA methods struggle to achieve positive transfer, our method consistently improves performance across both datasets and all three backbones. These promising results highlight not only the effectiveness of \ourmethod in real-world scenarios, but also underscore its versatility across various GAD architectures. 

\begin{table}[t!]
\centering
% \fontsize{9pt}{9pt}\selectfont
\resizebox{1\columnwidth}{!}{%
\begin{tabular}{lccc}
\toprule
\textbf{Model} & \textbf{Tolokers} & \textbf{Questions} & \textbf{Weibo} \\
\midrule
GCN & 63.20±0.26 & 51.58±0.44 & 86.30±14.06 \\
\quad + Ours & \textbf{66.27±0.07} & \textbf{60.61±0.97} & 96.16±0.01 \\
\quad \quad - Agg. Estimator & 66.18±0.01 & 42.95±0.01 & 96.16±0.01 \\
\quad \quad + Affinity Loss & 59.94±0.01 & 54.51±0.01 & \textbf{96.17±0.01} \\
\midrule
BWGNN & 62.02±0.87 & 50.40±0.45 & 93.04±1.10 \\
\quad + Ours & \textbf{65.51±0.04} & \textbf{63.21±0.37} & \textbf{95.26±0.49} \\
\quad  \quad - Agg. Estimator & 63.74±0.01 & 62.65±0.01 & 95.01±0.04 \\
\quad \quad  + Affinity Loss & 58.31±0.01 & 51.16±0.02 & 85.75±0.71 \\
\midrule
GHRN & 60.34±0.59 & 49.65±0.82 & 90.34±2.48 \\
\quad + Ours & \textbf{66.17±0.42} & \textbf{58.88±0.39} & 92.31±0.01 \\
\quad \quad  - Agg. Estimator & 64.54±0.01 & 58.05±0.01 & 92.28±0.01 \\
\quad \quad  + Affinity Loss & 56.71±0.02 & 55.24±0.06 & \textbf{92.32±0.01} \\
\bottomrule
\end{tabular}}
\caption{Performance (AUROC in \%) with standard deviation of \ourmethod and its variants.}
\label{table:ablation}
\end{table}

\noindent \textbf{Ablation Study. } To verify the contribution of key design in \ourmethod, we conduct ablation studies on the feature adapter and the alignment loss. To verify the effectiveness and robustness of our aggregation estimator module, we remove it in the variant (- Agg. Estimator). As shown in Table~\ref{table:ablation}, while the variant still achieves positive transfer on most datasets, its performance consistently drops compared to \ourmethod, highlighting the effectiveness of both the adapter module and dual branch design. Moreover, we further investigate the alignment loss by replacing it with an unsupervised GAD loss (+ Affinity Loss) as proposed in ~\cite{qiao2023truncated}. As Table~\ref{table:ablation} demonstrated, this variant leads to a performance reduction on all datasets except Weibo, where homophily dominates and GAD loss has limited additional benefit. In contrast, our dual-branch alignment loss does not rely on the homophily assumption of the graph, making it more suitable to mitigate the normality shift in GAD.

\begin{figure}[t!]
\centering

% Column titles
\begin{minipage}{0.05\linewidth}~
\end{minipage}%
\begin{minipage}{0.45\linewidth}
% \centering \textbf{Before}
\centering Before
\end{minipage}%
\begin{minipage}{0.45\linewidth}
% \centering \textbf{After}
\centering After
\end{minipage}

\vspace{0.5em}

% First row
\begin{minipage}{0.05\linewidth}
\centering
% \rotatebox[origin=c]{90}{\textbf{Photo}}
\rotatebox[origin=c]{90}{Photo}
\end{minipage}%
\begin{minipage}{0.47\linewidth}
\centering
\begin{subfigure}{\linewidth}
\includegraphics[width=\linewidth]{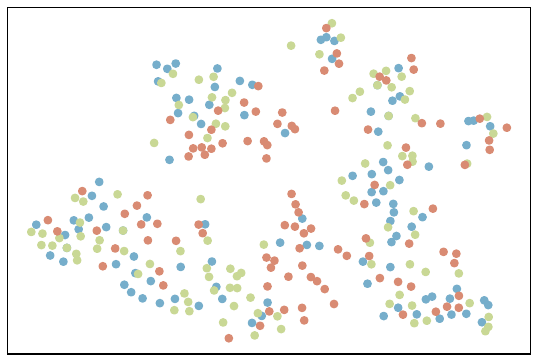}
\end{subfigure}
\end{minipage}%
\begin{minipage}{0.47\linewidth}
\centering
\begin{subfigure}{\linewidth}
    \includegraphics[width=\linewidth]{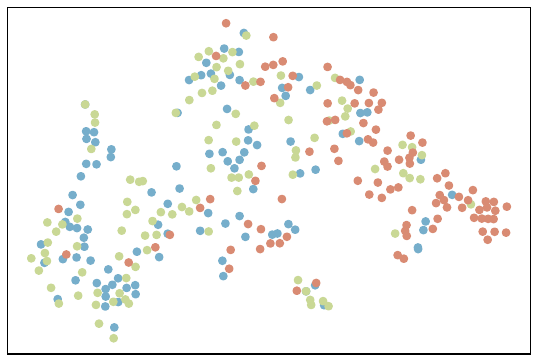}
\end{subfigure}
\end{minipage}
% Second row
\begin{minipage}{0.05\linewidth}
\centering
% \rotatebox[origin=c]{90}{\textbf{Computer}}
\rotatebox[origin=c]{90}{Computer}
\end{minipage}%
\begin{minipage}{0.47\linewidth}
\centering
\begin{subfigure}{\linewidth}
\includegraphics[width=\linewidth]{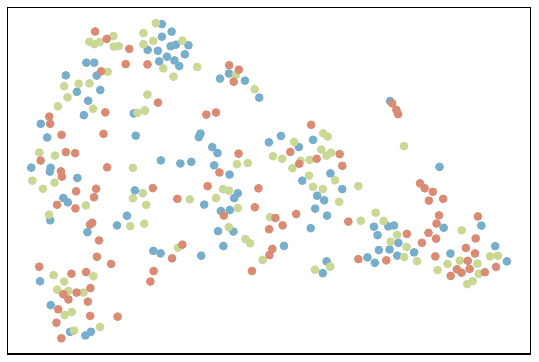}
\end{subfigure}
\end{minipage}%
\begin{minipage}{0.47\linewidth}
\centering
\begin{subfigure}{\linewidth}
\includegraphics[width=\linewidth]{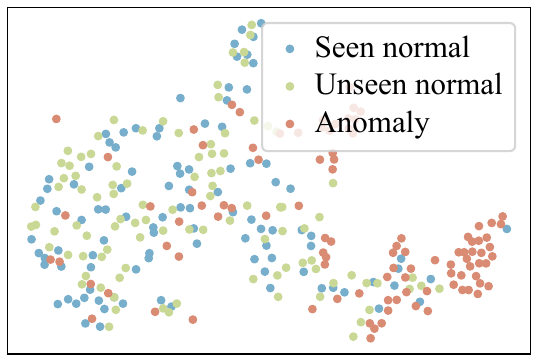}
\end{subfigure}
\end{minipage}

\caption{t-SNE visualizations of node representations by BWGNN (before and after applying \ourmethod) on two datasets.}
\label{fig:intro:comparison}

\end{figure}

\noindent \textbf{Case Study. }  To better understand why our proposed \ourmethod is more effective,  we visualize the representations before and after adaptation. As shown in Figure~\ref{fig:intro:comparison}, before adaptation, anomaly representations are indistinguishable from those of seen and unseen normals. This issue occurs because the pretrained GAD model fails to generalize to unseen normals, and their presence also affects the representations of seen normals through aggregation. With the help of \ourmethod, representations become much more discriminative, with anomalies clearly separated from all types of normal nodes. This is because our aligner is trained to mitigate the normality shift from the data level, helping the model preserve embedding quality and separability in the presence of unseen normals.

\section{Conclusion}
In this paper, we propose a novel TTA framework, \ourmethod, that can adapt to various GAD models to generalize to novel normal categories unseen during training, thereby guarding real-world graph applications under evolving normal definition. By utilizing the aggregation contamination as an indicator of normality shift, \ourmethod optimizes a graph aligner that operates at the graph level, making it compatible with a wide range of model architectures. Comprehensive empirical results show the effectiveness of \ourmethod in improving the generalizability of pre-trained GAD models compared to other graph TTA methods on 10 graph datasets against both artificial and real-world unseen normals.

\section{Acknowledgments}
This research was partly funded by the Australian Research Council (ARC)
under grant DP240101547 and the CSIRO – National Science Foundation (US) AI Research Collaboration Program under grant No. 62472416.

\bibliography{aaai2026}

\end{document}